\documentclass{article}

\usepackage[preprint]{neurips_2025}

\usepackage[utf8]{inputenc} 
\usepackage[T1]{fontenc}    
\usepackage{hyperref}       
\usepackage{url}            
\usepackage{booktabs}       
\usepackage{amsfonts}       
\usepackage{nicefrac}       
\usepackage{microtype}      
\usepackage{xcolor}         
\usepackage{graphicx}
\usepackage{amsmath}
\usepackage[table]{xcolor}
\usepackage{graphicx}
\usepackage{multirow}
\usepackage{float}
\usepackage{bbding}

\title{Towards Camera-Robust 3D Localization: Equation-Anchored Tool-Use for MLLMs}

\author{%
  Xueying Jiang$^{1}$, Wenhao Li$^{1}$, Quanhao Qian$^{2,3}$, Deli Zhao$^{2,3}$, \\
  \textbf{Shijian Lu}$^{1,}$\textsuperscript{\Envelope}, \textbf{Gongjie Zhang}$^{4,}$\textsuperscript{\Envelope}, \textbf{Ran Xu}$^{2,3,}$\textsuperscript{\Envelope} \\
  $^{1}$Nanyang Technological University \quad $^{2}$DAMO Academy, Alibaba Group \\
  $^{3}$HuPan Lab \quad $^{4}$Alibaba Group
}

\begin{document}

\maketitle
{\let\thefootnote\relax\footnotetext{\textsuperscript{\Envelope}Corresponding author.}}

\begin{abstract}

3D localization in Multimodal Large Language Models (MLLMs), including 3D object detection and 3D visual grounding, is fundamentally limited by camera intrinsic ambiguity: the same image admits different 3D scenes under different cameras. Existing MLLMs either ignore camera parameters and overfit to a canonical training intrinsic, or retrieve depth and 3D cues from external tools but treat the returned values as \emph{reference cues} (numerical hints that the model is free to interpret implicitly), both preventing camera information from being deterministically propagated into the prediction. We propose an \emph{equation-anchored tool-use} framework that re-purposes spatial tools as \emph{formula variables}. The proposed framework proactively retrieves camera intrinsics and samples multi-point metric depths, writes the pinhole back-projection equation $\hat{X} = (u_c - c_x)\bar{Z}/f_x$ explicitly in Chain-of-Thought (CoT), and substitutes tool outputs into the formula before regressing the final 9-DoF bounding box. On both 3D object detection and 3D visual grounding tasks under rescaled camera intrinsics from $0.5\times$ to $1.5\times$, our method outperforms RGB-only and tool-augmented baselines, with significant gains where the camera deviates most from the training scale. Code and data will be released.

\end{abstract}

\section{Introduction}

Recovering absolute 3D geometry from a single RGB image is a core capability for Multimodal Large Language Models to interact with the physical world, underpinning various 3D localization tasks such as 3D object detection and 3D visual grounding. While Multimodal Large Language Models (MLLMs) have achieved remarkable progress on 2D vision-language understanding~\cite{alayrac2022flamingo, bai2023qwen, liu2023visual, hurst2024gpt, yang2025qwen3, comanici2025gemini}, their performance on 3D localization tasks remains fundamentally limited. The reason is geometric: a single 2D image is inherently underdetermined, as the same projection admits infinitely many 3D scenes under different camera intrinsics.
Without an explicit mechanism to propagate camera parameters through reasoning, an MLLM is forced to rely on implicit statistical priors memorized from its training data. While scaling across diverse datasets may allow the model to approximate common camera configurations, this implicit approximation lacks a deterministic mathematical foundation, and its 3D predictions degrade sharply once the camera deviates from the familiar distribution. Camera-robust 3D localization, therefore, hinges on whether the model can deterministically propagate camera information through its reasoning chain rather than memorize a fixed intrinsic.

\begin{figure*}[t]
    \centering
    \includegraphics[width=1.0\linewidth]{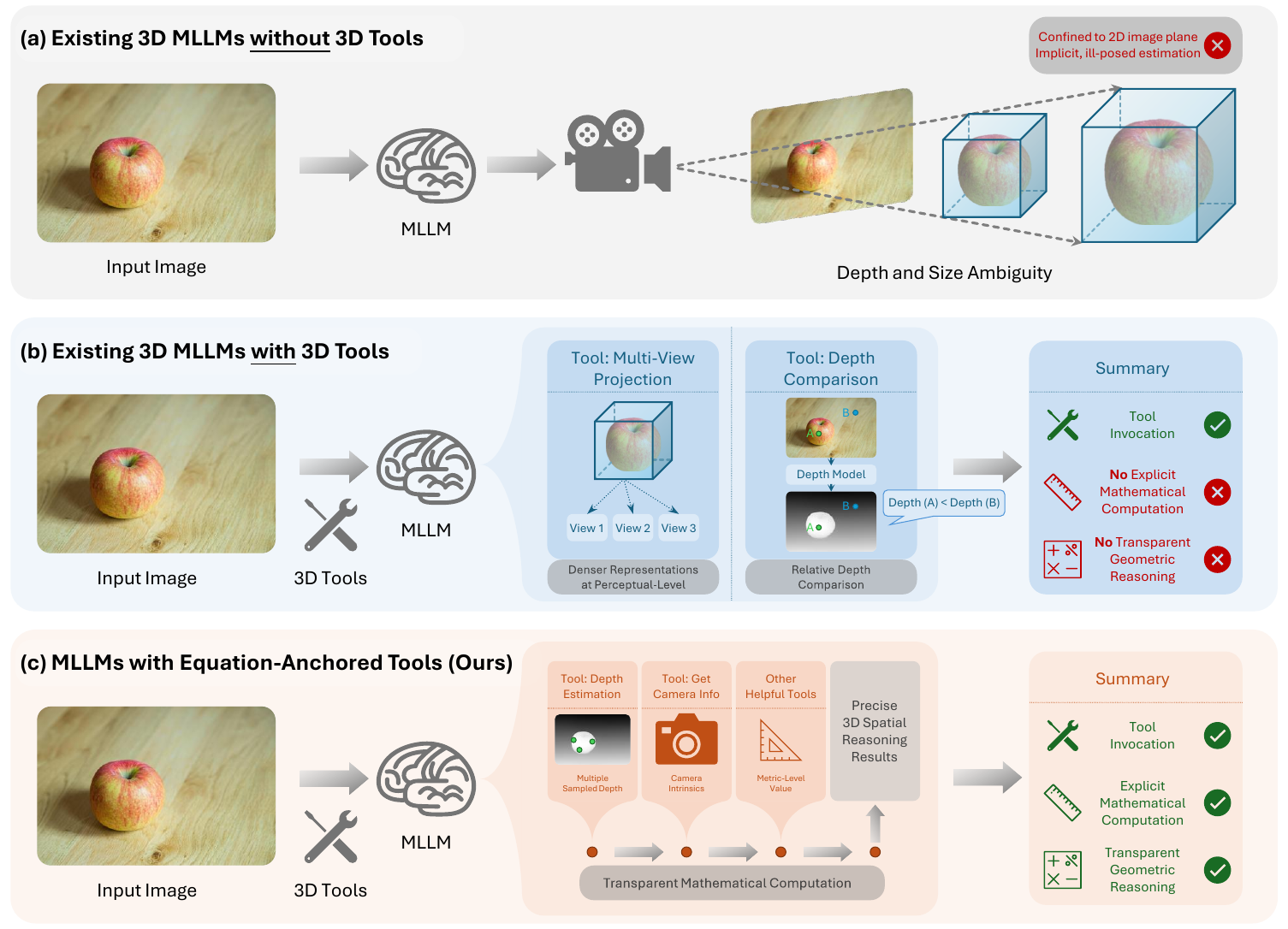}
    \caption{
    Comparison of spatial reasoning paradigms in Multimodal Large Language Models (MLLMs). (a) Existing RGB-only MLLMs rely on implicit spatial estimation confined to the 2D image plane, leading to depth and physical scale ambiguity. (b) Recent tool-augmented paradigm incorporates 3D tools but consumes the numerical returns merely as soft reference cues, failing to propagate camera parameters through the reasoning process. (c) Our proposed equation-anchored paradigm re-purposes retrieved metric depths and camera intrinsics as strict formula variables. By explicitly substituting these variables into the pinhole back-projection equation within a transparent Chain-of-Thought, the model deterministically deduces exact 3D bounding boxes, ensuring camera-robust spatial localization.
    }
    \label{fig:motivation}
\end{figure*}

As illustrated in Figure~\ref{fig:motivation}, two paradigms have been explored to address this challenge, yet both leave camera information weakly grounded in the reasoning process. 
Early RGB-only MLLMs~\cite{zheng2025video, zheng2025learning, zhang2025flatland} bypass camera parameters entirely and treat 3D localization as an end-to-end visual regression problem. Without access to camera intrinsics, they are forced to implicitly estimate physical scale based on visual cues derived from their training distribution. While this data-driven approximation may succeed under familiar camera setups, it lacks strict mathematical constraints, leaving the models vulnerable to geometric collapse when the camera undergoes out-of-distribution shifts.
More recently, tool-augmented MLLMs~\cite{zhou2025reinforced, wang2025last, tian2026last, zhang2026think3d} retrieve depth maps, metric distances, or even per-object 3D coordinates from external geometric tools. Although these methods successfully inject geometric information into the reasoning context, they consume the returned values as \emph{reference cues}: the MLLM is shown a numerical value, such as an object depth of 1.05m or a coordinate triplet, 
typically as a soft contextual hint rather than as the input to a specific geometric computation. For instance, Think3D~\cite{zhang2026think3d} leverages 3D tools primarily for multi-view rendering and reconstruction, treating the resulting views as denser visual context, while LAST~\cite{tian2026last} provides numerical per-object depths and absolute 3D coordinates that the MLLM is nevertheless free to ignore, approximate, or downgrade to relative hints during downstream reasoning. In all of these methods, regardless of whether the tool output is a depth map or an absolute coordinate, its \emph{usage pattern} stays implicit: the camera information 
is not constrained to flow through any specific geometric equation, and is therefore never deterministically propagated into the final prediction.

We argue that the central bottleneck of camera-robust 3D localization is not whether tool outputs are numerical or qualitative, but \emph{how the model is forced to use them}. Motivated by this, we propose an equation-anchored tool-use framework that re-purposes spatial tools as \emph{formula variables} rather than reference cues. Given a single RGB image, the MLLM proactively retrieves camera intrinsics $(f_x, f_y, c_x, c_y)$ via a Camera Intrinsic Tool, and samples multi-point metric depths within object masks via a Multi-Point Depth Sampling Tool. It writes the pinhole back-projection equation $\hat{X}=(u_c-c_x)\bar{Z}/f_x$, $\hat{Y}=(v_c-c_y)\bar{Z}/f_y$, $\hat{Z}=\bar{Z}$ explicitly in its Chain-of-Thought, substituting the retrieved tool outputs into the formula step by step before regressing the final 9-DoF bounding box. A dedicated training pipeline supervises the model on traces that interleave tool calls with such substitutions, turning the symbolic geometric chain into a fixed reasoning template.

This design bears two properties that structurally distinguish our framework from prior tool-augmented approaches. First, the camera intrinsics retrieved by the tool are no longer an optional context that the model may down-weight when visual heuristics suggest otherwise; they enter the prediction as multiplicative factors of the pinhole equation, so any change in $(f_x, f_y, c_x, c_y)$ deterministically rescales the predicted 3D center. This is precisely the property required for camera-robust 3D localization, allowing the model to generalize to camera configurations unseen during training without resorting to visual extrapolation. 
Second, by elevating tool invocation to an explicit equation-anchored computational loop, our design pioneers a mathematically-driven agentic paradigm for 3D localization. It enables MLLMs to autonomously invoke specialized geometric tools and process their numerical returns as strict mathematical variables rather than visual cues. This mathematically-grounded approach provides a starting point for future research in embodied AI, equipping spatial agents to actively interrogate complex physical environments instead of simple visual approximations.

Extensive experiments validate the efficacy of this paradigm. We benchmark our framework on 3D object detection and 3D visual grounding tasks, evaluating under varying camera rescale factors. Results show that existing baselines suffer from geometric collapse when the camera deviates from the familiar training scale. In contrast, by interleaving tool retrieval with transparent symbolic substitution to deduce exact 3D coordinates, our equation-anchored framework exhibits camera robustness, effectively achieving better spatial localization performance across all tasks.

The major contributions of this work are threefold. First, we identify the distinction between \emph{reference cues} and \emph{formula variables} as the core lever for camera-robust 3D localization in MLLMs: existing tool-augmented methods provide camera-relevant numerical values but consume them as reference cues, which prevents deterministic camera propagation regardless of whether those values are relative or absolute. Second, we propose an equation-anchored tool-use framework that forces the MLLM to substitute retrieved camera intrinsics and multi-point depths into an explicit pinhole equation during Chain-of-Thought, before regressing the final 9-DoF 3D bounding box. The resulting reasoning chain is auditable, exposing every intermediate quantity for failure analysis. Third, we evaluate on both 3D object detection and 3D visual grounding tasks under rescaled camera intrinsics from $0.5\times$ to $1.5\times$, where our method outperforms both RGB-only and tool-augmented baselines, with significant gains appearing precisely where the camera deviates most from the training scale.

\section{Related Work}
\subsection{MLLMs for Spatial Intelligence}
While Multimodal Large Language Models have demonstrated remarkable success in standard 2D vision-language tasks~\cite{alayrac2022flamingo, bai2023qwen, liu2023visual, hurst2024gpt, yang2025qwen3, comanici2025gemini}, they inherently struggle with 3D spatial intelligence~\cite{yang2025thinking, zheng2025learning, chen2024ll3da, yu2025far, cai2025scaling, fan2025vlm, wu2025spatial, feng2025survey, zhang2026generalization}, which requires a precise comprehension of physical scales, geometry, and spatial relationships. To mitigate this limitation, many prior works have focused on enriching spatial representations by incorporating auxiliary cues, such as depth maps~\cite{cheng2024spatialrgpt, zhou2025roborefer, liu2025ssr} and segmentation masks~\cite{wang2023chat, huang2024chat}. However, most of these approaches treat auxiliary data as enhanced perceptual features rather than explicit variables for precise spatial reasoning. Alternatively, models tackling 3D scene understanding with RGB-only inputs~\cite{zheng2025video, zheng2025learning, zhang2025flatland} attempt to interpret 3D environments from monocular visual cues, but inevitably suffer from geometric collapse and depth ambiguity due to their restriction to the 2D image plane. To resolve this, our paradigm bridges abstract visual perception with concrete geometric logic, transforming spatial understanding from implicit estimation into explicit mathematical reasoning.

\subsection{Tool-Assisted Reasoning}
To extend the reasoning capabilities of foundation models, early tool-augmented reasoning frameworks~\cite{nakano2021webgpt, shen2023hugginggpt, wu2023visual} empowered LLMs to invoke external tools, such as code generation~\cite{suris2023vipergpt, gao2023pal} and symbolic solvers~\cite{pan2023logic, li2024neuro}, enabling them to tackle complex logical and mathematical problems beyond their innate capabilities. 
While these text-based reasoning frameworks excel in logic and mathematics, they fundamentally fail to comprehend continuous spatial environments due to a lack of visual engagement. To bridge this gap, ``thinking with images'' paradigms~\cite{su2025thinking, huang2026vision, zhang2026thyme, zheng2026deepeyes} have been proposed.
In these multimodal Chain-of-Thought frameworks, some MLLMs~\cite{hu2024visual} invoke specialized 2D vision experts, such as object tracking, grounding, and segmentation models, to generate intermediate visual evidence, while others perform intuitive 2D-level visual operations like cropping or masking~\cite{fu2025refocus, su2025openthinkimg, wang2025pixel, sarch2025grounded, zheng2026deepeyes}. 

To enrich the spatial context for complex 3D scene comprehension, recent efforts incorporate specialized spatial vision tools for depth estimation~\cite{zhou2025reinforced, wang2025last, tian2026last} and 3D reconstruction~\cite{zhou2025reinforced, wang2025last, tian2026last}.
However, since dense tool outputs like high-resolution depth maps are difficult for MLLMs to directly interpret, most existing paradigms utilize them merely as relative visual cues. Rather than treating these outputs as coarse visual references, our method explicitly formulates spatial tool outputs as mathematical variables. By integrating multi-point depth samples and camera intrinsics into an explicit geometric reasoning chain, our agent performs transparent mathematical calculations, grounding 3D localization in explicit geometric formulations rather than implicit spatial regression.

\begin{figure*}[t]
    \centering
    \includegraphics[width=1.0\linewidth]{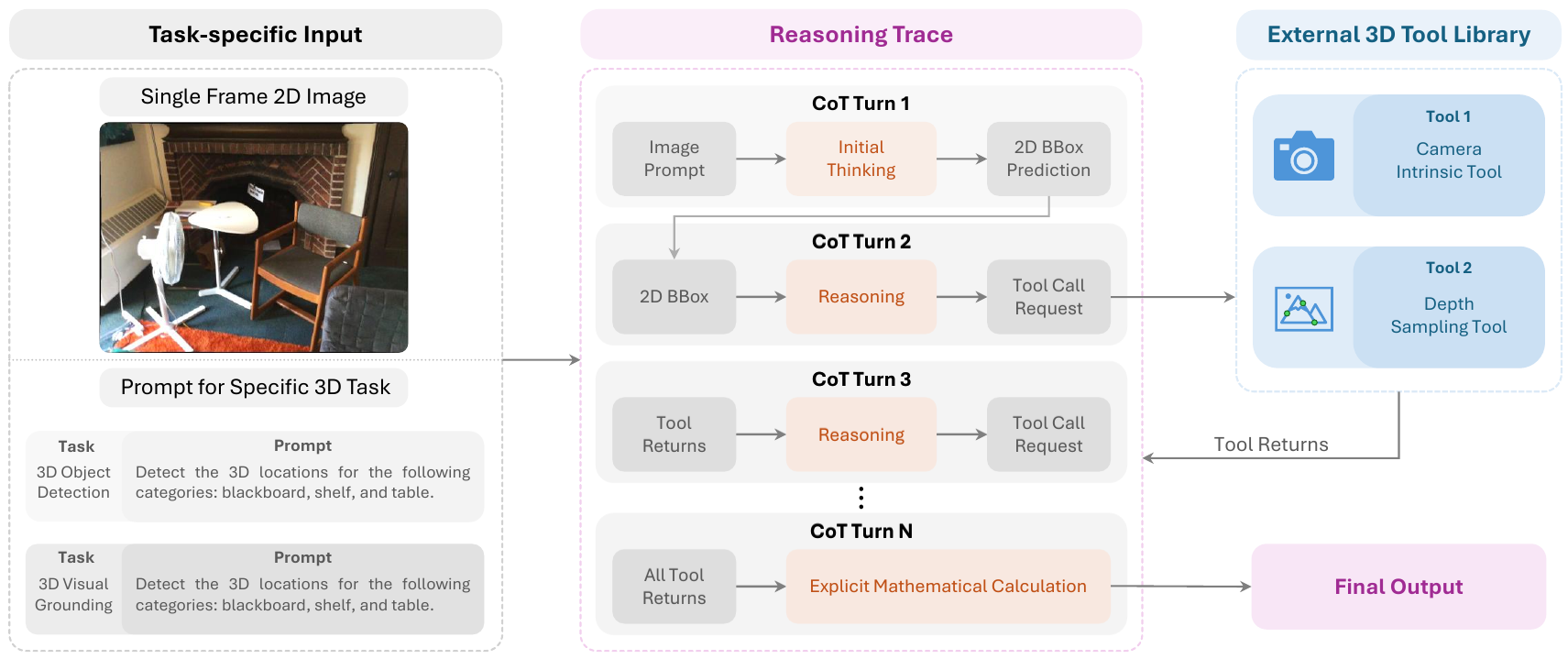}
    \caption{
    Overview of our proposed equation-anchored spatial agent. Given a single-frame RGB image and a task-specific prompt, the agent first leverages its internal visual perception to predict initial 2D bounding boxes without relying on external detectors. Subsequently, it engages in a structured multi-turn Chain-of-Thought reasoning trace, utilizing these 2D spatial anchors to interactively retrieve exact camera parameters and metric depths from an external 3D Tool Library. In the concluding turn, rather than treating the tool outputs as soft reference cues, the agent strictly re-purposes them as formula variables. By explicitly substituting these variables into the pinhole back-projection equation, the model deterministically deduces exact 3D coordinates, finally yielding camera-robust 9-DoF 3D localizations.
    }
    \label{fig:overall_architecture}
\end{figure*}

\section{Method}
\subsection{Overall Architecture}

Figure~\ref{fig:overall_architecture} illustrates the overall architecture of our proposed equation-anchored spatial agent. To achieve camera-robust 3D spatial understanding and prediction, we empower the Multimodal Large Language Model (MLLM) to interactively acquire explicit geometric data via a structured tool-use-and-reasoning loop.

Given a single-frame RGB image $I$ and a task-specific natural language prompt $T$, the agent initiates the first Chain-of-Thought turn. It leverages its internal visual perception to predict a set of 2D bounding boxes $\mathcal{B}_{2D}$ for the target objects directly, without relying on external 2D detectors. Subsequently, the agent utilizes these 2D bounding boxes as initial spatial anchors to interact with an external 3D Tool Library. Over successive CoT turns, the MLLM issues tool call requests, such as the Camera Intrinsic Tool and the Depth Sampling Tool, to retrieve precise camera parameters $(f_x, f_y, c_x, c_y)$ and explicit metric depth observations $(u, v, Z)$ within the specified 2D regions. 
Importantly, rather than treating these retrieved values as implicit \emph{reference cues}, our framework strictly re-purposes them as \emph{formula variables}. By substituting the retrieved depths and camera intrinsics directly into the pinhole back-projection equation within its reasoning chain, the agent deterministically deduces the exact 3D spatial center $(X_c, Y_c, Z_c)$. In the concluding CoT turn, grounded by this explicit geometric computation, the model outputs the final task-specific results formatted as a JSON list of 9-DoF 3D bounding boxes $\mathcal{B}_{3D}$. Each bounding box is represented as ${bbox}_{3D} = [X_c, Y_c, Z_c, l, w, h, \psi, \theta, \phi]$, explicitly parameterizing the 3D center, physical dimensions, and orientation angles in the camera coordinate system.

\subsection{3D Spatial Tools}

To realize our equation-anchored framework, we introduce a geometry-driven 3D spatial toolkit designed to extract strict formula variables rather than implicit perceptual hints. This toolkit equips the MLLM with explicit and numerical geometric measurements, serving as the rigorous mathematical foundation for precise 3D spatial grounding.

\textbf{Multi-Point Depth Sampling.} To acquire object-level geometric data, the model queries the Depth Sampling Tool with a list of target objects. Each target object is formulated as a semantic category label $c$ paired with a 2D bounding box ${bbox}_{2D}$, formalized as a query tuple $(c, {bbox}_{2D})$, where ${bbox}_{2D}=[u_{\min}, v_{\min}, u_{\max}, v_{\max}]$. The tool leverages a promptable segmentation model~\cite{carion2025sam} to extract a precise object mask based on the initial 2D spatial anchor. From this masked region, it samples multiple valid pixels (up to $N$ points per object) and returns depth triplets:
\begin{equation}[(u_1,v_1,Z_1),\dots,(u_N,v_N,Z_N)],\end{equation}
where $(u_i, v_i)$ represents the 2D pixel coordinates of the sampled point, $Z_i$ denotes the corresponding metric depth in meters, and $N$ is set to 5. Specifically, the metric depth values $Z_i$ are directly extracted from a dense depth map provided by datasets or predicted by a monocular depth estimator~\cite{piccinelli2025unidepthv2} conditioned on the retrieved intrinsics. By explicitly formatting these outputs as exact numerical triplets, the tool provides formula variables necessary for the MLLM to deterministically compute real-world physical scale and object localization.

\textbf{Camera Intrinsic Retrieval.} Given the image context, the Camera Intrinsic Tool extracts the focal lengths and principal point coordinates $(f_x, f_y, c_x, c_y)$ from the dataset-provided camera calibration parameters. Importantly, rather than feeding these parameters into the MLLM as soft contextual tokens, our framework strictly mandates their use as multiplicative factors in the back-projection equation. These parameters enable the explicit computation of exact 3D spatial locations from sampled pixel coordinates $(u, v)$ and metric depth $Z$, thereby mitigating scale ambiguity and guaranteeing camera-robust generalization.

\subsection{Explicit Geometric Chain-of-Thought}

We formulate an explicit Geometric Chain-of-Thought to achieve camera-robust metric 3D localization. By forcing the model to explicitly substitute retrieved tool outputs as formula variables into mathematical equations rather than treating them as soft reference cues, we enforce a transparent and interpretable reasoning trajectory spanning five sequential steps:

\textbf{Step 1: Semantic Grounding on the 2D Plane.} Leveraging the robust 2D perception capabilities inherent in MLLMs, the agent directly localizes target objects on the 2D image plane without relying on external detectors. It initially predicts bounding boxes in a normalized coordinate space, which aligns with the native localization format of the MLLMs' original training paradigm. To seamlessly bridge these intermediate 2D predictions with subsequent geometric tool invocations, the reasoning trajectory then forces the model to calculate the absolute pixel coordinates:
\begin{equation}
u_{\text{abs}}=\text{round}\left(\frac{u_{\text{norm}}}{1000}\cdot W\right),\quad v_{\text{abs}}=\text{round}\left(\frac{v_{\text{norm}}}{1000}\cdot H\right),
\end{equation}
where $u_{\text{norm}}$ and $v_{\text{norm}}$ represent normalized coordinates, $u_{\text{abs}}$ and $v_{\text{abs}}$ represent absolute coordinates, and $W$ and $H$ denote the image width and height.

\textbf{Step 2: Camera Intrinsic Retrieval.} To initiate the 2D-to-3D spatial lifting, the model first issues a tool call to retrieve the camera parameters. This step isolates the camera configuration from implicit visual features, extracting the exact focal lengths and principal point coordinates $(f_x,f_y,c_x,c_y)$ required as strict formula variables for subsequent back-projection.

\textbf{Step 3: Metric Depth Acquisition.} Subsequently, the agent invokes the Depth Sampling Tool to acquire the depth of the target objects from multiple points. By querying the tool with the localized 2D bounding boxes, the model collects valid metric depth values within the object-specific segmentation masks. To ensure geometric reliability, noisy depths smaller than $0.1$ meters are systematically discarded from the tool's response, yielding a robust set of depth variables.

\textbf{Step 4: Rigorous Mathematical Deduction.} Rather than consuming the tool outputs as implicit reference cues, the Chain-of-Thought executes an equation-anchored deduction. Given the predicted absolute 2D bounding box derived in Step 1, denoted as ${bbox}_{2D}=[u_{\min},v_{\min},u_{\max},v_{\max}]$, the model explicitly computes the 2D bounding box center via:
\begin{equation}
u_c = \frac{u_{\min} + u_{\max}}{2}, \quad v_c = \frac{v_{\min} + v_{\max}}{2}. 
\end{equation}

The sampled depths $\{Z_i\}_{i=1}^N$ are aggregated into an average depth $\bar{Z}$. Using the pinhole camera model, the model substitutes these extracted variables to deduce the initial 3D center:
\begin{equation}
\hat{X}=\frac{(u_c-c_x)\cdot\bar{Z}}{f_x},\quad\hat{Y}=\frac{(v_c-c_y)\cdot\bar{Z}}{f_y},\quad\hat{Z}=\bar{Z}. 
\end{equation}

This step structurally guarantees camera-robustness, explicitly forcing any change in the retrieved camera intrinsics to deterministically rescale the deduced 3D center.

\textbf{Step 5: Geometry-Conditioned 3D Spatial Estimation.} Finally, the agent anchors its 3D bounding box regression on the explicitly computed geometric center. It consolidates this deterministic mathematical context with visual cues to reason and output the precise 9-DoF bounding box ${bbox}_{3D}$:
\begin{equation}
{bbox}_{3D}=[X_c,Y_c,Z_c,l,w,h,\psi,\theta,\phi].
\end{equation}

By deterministically propagating camera information through this explicit mathematical chain, the model overcomes the intrinsic ambiguity inherent in purely visual estimations, effectively locking the final 3D prediction to the exact camera scale.

\subsection{Training Pipeline}
To empower standard MLLMs with this explicit 3D reasoning capability, we design a geometry-guided supervised fine-tuning (SFT) pipeline, where the model is supervised using structured, multi-turn equation-anchored reasoning traces rather than direct end-to-end 3D bounding box annotations.

\textbf{Data Construction.} We construct rigorously structured reasoning data based on the ScanNet~\cite{dai2017scannet} dataset, which features diverse indoor scenes to support the 3D object detection task. For 3D visual grounding, we process target frames from the ScanRefer~\cite{chen2020scanrefer} dataset to facilitate the construction of single-frame reasoning samples, utilizing Qwen3.5 to generate and verify context-aware referring expressions. For both tasks, ground-truth 3D boxes are projected onto the image plane to obtain 2D regions. Based on the 2D regions and object categories, we then employ SAM3~\cite{carion2025sam} to generate precise object masks, from which we extract point-wise metric depth values to simulate the exact outputs of the Multi-Point Depth Sampling tool. Importantly, these trajectories strictly format ground-truth geometric measurements as explicit formula variables rather than reference cues, embedding them into a standardized multi-turn template (\texttt{<tool\_call>}, \texttt{<tool\_response>}, \texttt{<think>}, \texttt{<answer>}). This structural constraint ensures the model rigorously internalizes the deterministic, step-by-step geometric derivation.

\textbf{Model Training.} We adopt Qwen3-VL-8B-Instruct~\cite{yang2025qwen3} as our base model, optimizing it via a masked supervised causal language modeling objective. To preserve pretrained visual representations, the visual encoder and projector are frozen, updating only the decoder and language modeling head. Within chat templates, user instructions and system prompts are masked from the loss, while optimization is applied to assistant-side tokens. Supervising the model on these equation-anchored tool-use traces forces the LLM to anchor its spatial reasoning in deterministic mathematics.

\begin{table*}[t]
    \centering
    \caption{3D Object Detection performance across rescale factors. Best in \textbf{bold}, second \underline{underlined}.}
    \label{tab:3d_detection_rescale}
    \resizebox{1.0\linewidth}{!}{
    \begin{tabular}{lccccccccccc}
        \toprule
        Method & 0.5 & 0.6 & 0.7 & 0.8 & 0.9 & 1.0 & 1.1 & 1.2 & 1.3 & 1.4 & 1.5 \\
        \midrule
        \multicolumn{12}{l}{\cellcolor{gray!20}\textit{Generalist MLLMs}} \\
        GPT-5.4-2026-03-05~\cite{openai2026gpt54} & 2.67  & 1.11 & 0.81 & 1.96  & 0.83  & 0.63 & 0.41  & 1.37  & 1.31  & 1.34 & 0.59  \\
        DeepSeek-v4-Flash~\cite{deepseek2026v4} & 0.00 & 0.24 & 0.00 & 0.11 & 0.00 & 0.12 & 0.00 & 0.17  & 0.00 & 0.42 & 0.00 \\
        Claude-Sonnet-4.6~\cite{anthropic2026sonnet46} & 0.87 & 1.50 & 1.35 & 0.79 & 0.53 & 0.94  & 1.64 & 0.78 & 2.44  & 2.70  & 1.84  \\
        Claude-Opus-4.7~\cite{anthropic2026opus47} & 0.00 & 0.00 & 0.00 & 0.00 & 0.00 & 0.00 & 0.00 & 0.00 & 0.00 & 0.00 & 0.00 \\
        Kimi-K2.6~\cite{kimi2026k26} & 0.28 & 0.00 & 0.06 & 0.08 & 0.08 & 0.00  & 0.17 & 0.08 & 0.08 & 0.00 & 0.00 \\
        Qwen3.6-Flash~\cite{qwen2026q36flash} & 4.03 & 9.41 & 10.01 & 19.57 & 18.94 & 25.68 & 19.47 & 16.82  & 19.94 & 17.40  &  17.88 \\
        Seed-2.0-Pro-260215~\cite{doubao2026seed2pro} & 2.00  & 5.58  & 7.56  & 14.50  & 17.40  & 24.00  & 20.84 & 19.23  & 17.20  & 14.28 & 12.54 \\
        \midrule
        \multicolumn{12}{l}{\cellcolor{gray!20}\textit{Specialist MLLMs}} \\
        VG LLM-8B~\cite{zheng2025learning} & 13.19 & 16.77 & 24.30 & 25.12 & 29.67  & 38.49 & 32.78  & 30.40 & 23.08  & \underline{24.07}  & 17.16 \\
        Qwen3-VL-8B-Instruct~\cite{yang2025qwen3} & 10.75 & 13.83 & 23.99 & 33.15 & 33.29 & 38.20 & 31.18 & 27.17 & 22.04 & 18.52 & 15.05 \\
        Ours-8B (w/ UniDepthV2~\cite{piccinelli2025unidepthv2}) & \underline{16.97} & \textbf{30.85} & \underline{33.33} & \underline{35.26} & \textbf{42.38} & \underline{44.13} & \underline{39.70} & \underline{31.88} & \underline{27.16} & 21.65 & \underline{20.76} \\
        Ours-8B (w/ GT depth) & \textbf{18.93} & \underline{30.03} & \textbf{36.08} & \textbf{37.42} & \underline{41.69} & \textbf{44.90} & \textbf{39.81} & \textbf{34.25} & \textbf{28.48} & \textbf{24.40} & \textbf{24.29} \\
        \bottomrule
    \end{tabular}
    }
\end{table*}

\begin{table*}[t]
    \centering
    \caption{3D Visual Grounding performance across rescale factors. Best in \textbf{bold}, second \underline{underlined}.}
    \label{tab:3d_grounding_rescale}
    \resizebox{1.0\linewidth}{!}{
    \begin{tabular}{lccccccccccc}
        \toprule
        Method & 0.5 & 0.6 & 0.7 & 0.8 & 0.9 & 1.0 & 1.1 & 1.2 & 1.3 & 1.4 & 1.5 \\
        \midrule
        \multicolumn{12}{l}{\cellcolor{gray!20}\textit{Generalist MLLMs}} \\
        GPT-5.4-2026-03-05~\cite{openai2026gpt54} & 0.00 & 0.00 & 0.00 & 0.00 & 0.00 & 0.00 & 0.00 & 0.00 & 0.00 & 0.00 & 0.00 \\
        DeepSeek-v4-Flash~\cite{deepseek2026v4} & 0.53 & 0.27 & 0.27 & 0.00 & 0.53 & 0.53 & 0.80 & 0.27 & 0.80 & 0.27 & 0.27 \\
        Claude-Sonnet-4.6~\cite{anthropic2026sonnet46} & 0.27 & 0.00 & 0.00 & 0.00 & 0.00 & 0.00 & 0.27 & 0.00 & 0.00 & 0.00 & 0.27 \\
        Claude-Opus-4.7~\cite{anthropic2026opus47} & 0.00 & 0.00 & 0.00 & 0.00 & 0.00 & 0.00 & 0.00 & 0.00 & 0.00 & 0.00 & 0.00 \\
        Kimi-K2.6~\cite{kimi2026k26} & 0.00 & 0.00 & 0.00 & 0.00 & 0.00 & 0.00 & 0.00 & 0.00 & 0.00 & 0.00 & 0.00 \\
        Qwen3.6-Flash~\cite{qwen2026q36flash} & 3.71 & 7.96 & 11.67 & 19.89 & 26.79 & 33.42 & 40.05 & 38.46 & 31.30 & 25.20 & 22.28 \\
        Seed-2.0-Pro-260215~\cite{doubao2026seed2pro} & 4.51 & 9.28 & 17.24 & 28.65 & 36.34 & 40.58 & 32.63 & 25.46 & 20.95 & 12.73 & 10.88 \\
        \midrule
        \multicolumn{12}{l}{\cellcolor{gray!20}\textit{Specialist MLLMs}} \\
        VG LLM-8B~\cite{zheng2025learning} & \textbf{17.51} & 22.28 & 27.06 & 34.75 & 43.77 & 48.28 & 47.75 & 40.85 & 35.28 & 30.77 & 24.67 \\
        Qwen3-VL-8B-Instruct~\cite{yang2025qwen3} & 8.49 & 16.98 & 28.38 & 41.64 & \underline{51.72} & 53.85 & 48.54 & 40.32 & 32.36 & 23.61 & 20.69 \\
        Ours-8B (w/ UniDepthV2~\cite{piccinelli2025unidepthv2}) & \underline{13.26} & \underline{24.14} & \underline{38.20} & \underline{48.28} & \textbf{57.03} & \underline{55.17} & \underline{57.82} & \underline{49.60} & \underline{43.50} & \underline{35.28} & \underline{28.12} \\
        Ours-8B (w/ GT depth) & 11.41 & \textbf{25.46} & \textbf{42.71} & \textbf{50.13} & \textbf{57.03} & \textbf{58.62} & \textbf{58.89} & \textbf{51.19} & \textbf{43.77} & \textbf{36.07} & \textbf{32.89} \\
        \bottomrule
    \end{tabular}
    }
\end{table*}

\section{Experiments}

\subsection{Experimental Settings}

\textbf{Evaluation Metrics.} For 3D object detection, we adopt the Average F1 (Avg F1) score with an Intersection over Union (IoU) threshold of 0.25 as the evaluation metric. For 3D visual grounding, we report the accuracy at an IoU threshold of 0.25 (Acc@IoU=0.25). Unless otherwise specified, the reported performance for 3D object detection is averaged across 31 common categories.

\begin{table}[t]
    \centering
    \caption{Ablation study of different spatial tools (rescale factor = 1.0).}
    \label{tab:ablation_spatial_tools}
    \setlength{\tabcolsep}{5.5pt}
    \scalebox{0.9}{
    \begin{tabular}{cccccc}
        \toprule
        \multicolumn{2}{c}{Spatial Tools} & \multicolumn{3}{c}{3D Object Detection} & 3D Visual Grounding \\
        \cmidrule(r){1-2} \cmidrule(lr){3-5} \cmidrule(l){6-6}
        Camera Intrinsic & Depth Sampling & Category 8 & Category 20 & Category 31 & Performance \\
        \midrule
                   &            & 60.73 & 45.90 & 38.43 & 50.13 \\
        \checkmark &            & 63.11 & 48.38 & 40.27 & 51.46 \\
                   & \checkmark & 64.63 & 48.29 & 40.09 & 54.38 \\
        \checkmark & \checkmark & \textbf{67.47} & \textbf{54.05} & \textbf{44.13} & \textbf{55.17} \\
        \bottomrule
    \end{tabular}
    }
\end{table}

\begin{table}[t]
    \centering
    \caption{Ablation study of different depth sources in the reasoning chain (rescale factor = 1.0).}
    \label{tab:ablation_depth_source}
    \setlength{\tabcolsep}{13pt}
    \scalebox{0.9}{
    \begin{tabular}{lcccc}
        \toprule
         & \multicolumn{3}{c}{3D Object Detection} & 3D Visual Grounding \\
        \cmidrule(lr){2-4} \cmidrule(l){5-5}
        Depth Source & Category 8 & Category 20 & Category 31 & Performance \\
        \midrule
        UniDepthV2~\cite{piccinelli2025unidepthv2} & 67.47 & 54.05 & 44.13 & 55.17 \\
        GT Depth   & \textbf{70.51} & \textbf{55.27} & \textbf{44.90} & \textbf{58.62} \\
        \bottomrule
    \end{tabular}
    }
\end{table}

\begin{table}[!t]
    \centering
    \caption{Ablation study of the explicit equation-anchored CoT reasoning (rescale factor = 1.0).}
    \label{tab:ablation_cot}
    \setlength{\tabcolsep}{15pt}
    \scalebox{0.9}{
    \begin{tabular}{ccccc}
        \toprule
         & \multicolumn{3}{c}{3D Object Detection} & 3D Visual Grounding \\
        \cmidrule(lr){2-4} \cmidrule(l){5-5}
        Explicit CoT & Category 8 & Category 20 & Category 31 & Performance \\
        \midrule
                   & 67.21 & 47.66 & 38.20 & 53.85 \\
        \checkmark & \textbf{67.47} & \textbf{54.05} & \textbf{44.13} & \textbf{55.17} \\
        \bottomrule
    \end{tabular}
    }
\end{table}

\subsection{Benchmarking with the State-of-the-Art}

\textbf{Quantitative Benchmarking.} Tables~\ref{tab:3d_detection_rescale} and~\ref{tab:3d_grounding_rescale} present 3D object detection and 3D visual grounding results under varying rescale factors. Compared to specialist MLLMs, our framework achieves superior performance and better camera-robustness. The superior performance is largely attributed to two key design choices. First, rather than treating tool outputs as soft cues, our model extracts metric depths and intrinsics strictly as formula variables. Second, explicit equation-anchored reasoning deterministically propagates these variables through the back-projection equation for rigorous 3D estimation. Consequently, RGB-only models like VG LLM and Qwen3-VL-8B-Instruct overfit to canonical training intrinsics and suffer geometric collapse during camera shifts.
For instance, in the 3D visual grounding task, the performance of Qwen3-VL-8B-Instruct drops from 53.85\% at the original scale to 8.49\% at the 0.5$\times$ scale.
Furthermore, our framework equipped with ground-truth (GT) depth outperforms the variant using UniDepthV2 in most scales. This performance gain shows that with more accurate intermediate variables, our strict mathematical deduction chain can achieve better 3D localization performance.

Moreover, evaluations of generalist MLLMs reveal a systematic failure in absolute 3D localization across varying camera scales. Despite their extensive multi-modal pre-training, these models yield near-zero performance as they lack the explicit geometric constraints required to resolve depth and scale ambiguity. In the absence of camera intrinsics, absolute metric regression becomes a mathematically ill-posed problem, where even minor discrepancies in coordinate estimation result in zero IoU under the stringent 3D evaluation criteria. This degradation confirms that 3D spatial localization requires a transition from implicit heuristic estimation to a deterministic, equation-anchored reasoning paradigm.

\subsection{Ablation Study}

\textbf{Effectiveness of Spatial Tools.} As Table~\ref{tab:ablation_spatial_tools} shows, the baseline lacking both spatial tools struggles across all evaluation metrics. Relying on purely implicit 2D-to-3D projection, it suffers from severe scale ambiguity. Incorporating the Camera Intrinsic Tool yields clear improvements by establishing the explicit geometric relationship between 2D pixels and 3D projection rays. Similarly, introducing the Depth Sampling Tool provides improvements by equipping the model with metric depth values to break the depth ambiguity. Finally, the best performance is achieved when both spatial tools are included. This combination successfully completes the required variables for the pinhole back-projection equation, effectively elevating the MLLM from a perceptual estimator to a mathematically-grounded spatial agent.

\textbf{Impact of Depth Source Quality.} In Table~\ref{tab:ablation_depth_source}, we investigate the effect of depth source in our framework by substituting the predicted depths (UniDepthV2) with Ground Truth (GT) depths within the reasoning chain. Equipping the tool with GT depth yields substantial improvements, notably boosting 3D visual grounding from 55.17\% to 58.62\% and 8-category 3D object detection from 67.47\% to 70.51\%. Since our framework enforces a strict mathematical propagation of tool outputs, its geometric correctness is directly affected by the perception quality of the external tool.

\textbf{Necessity of Explicit CoT Reasoning.} Table~\ref{tab:ablation_cot} highlights the critical role of explicit mathematical derivation. Removing the equation-anchored Chain-of-Thought (w/o CoT) forces the model to directly regress 3D bounding boxes. This opaque mapping causes severe performance drops in complex scenarios (e.g., 31-category performance falls from 44.13\% to 38.20\%). In contrast, explicit CoT ensures the transparent substitution of geometric variables into projection formulas. This performance gap demonstrates that skipping rigorous mathematical reasoning downgrades the model to unreliable implicit guessing, and equation-driven deduction is essential for robust 3D understanding.

\begin{figure*}[t]
    \centering
    \includegraphics[width=0.99\linewidth, height=297pt]{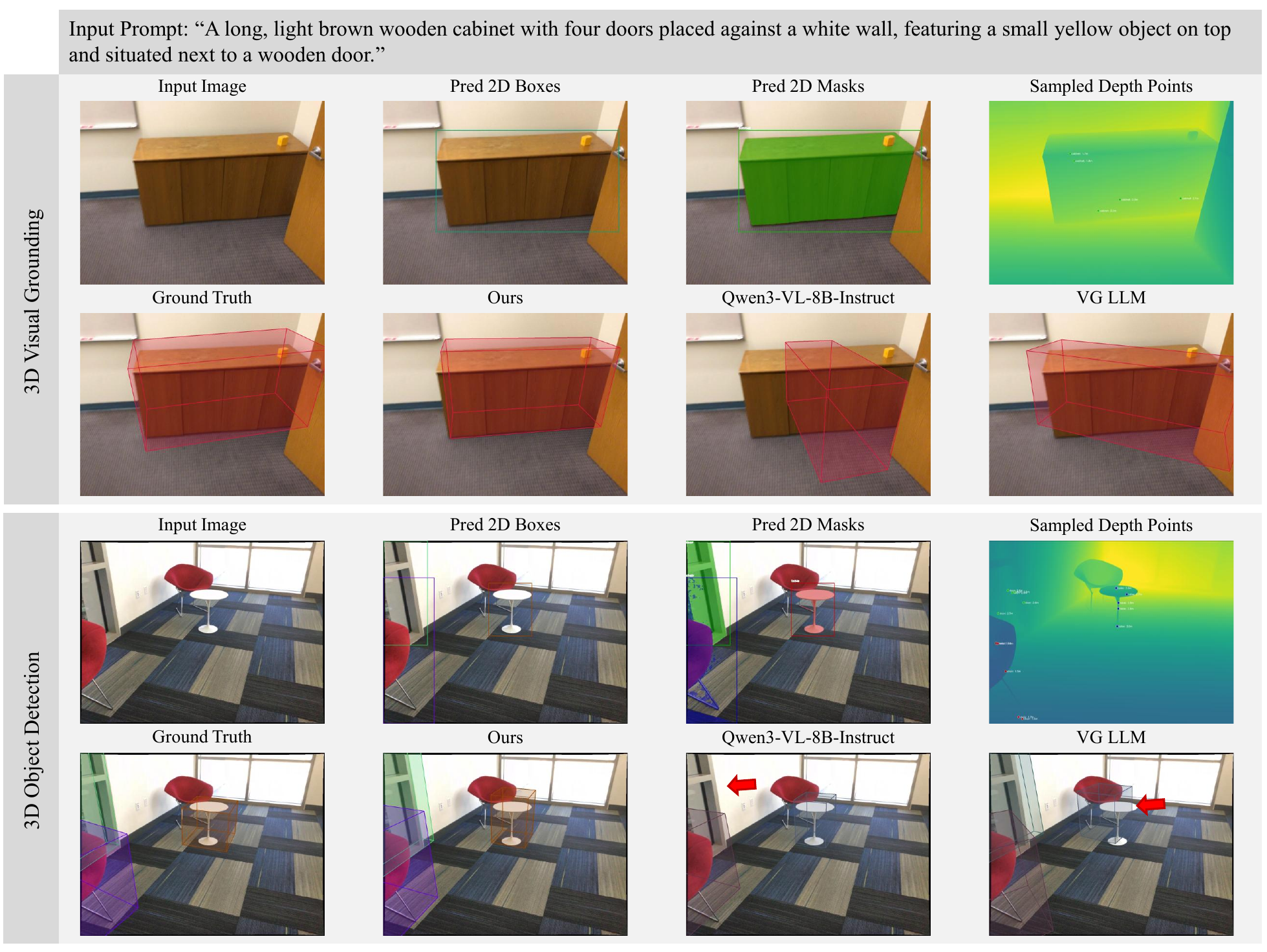}
    \caption{Qualitative comparison on 3D visual grounding (top) and 3D object detection (bottom).  Red arrows highlight incorrect predictions.}
    \label{fig:visualization_results}
\end{figure*}

\subsection{Visualization Results}

Figure~\ref{fig:visualization_results} provides qualitative results on both the 3D visual grounding (top) and 3D object detection (bottom) tasks. We demonstrate the intermediate spatial tool outputs of our framework, including the predicted 2D boxes, 2D masks, and sampled depth points, to illustrate the transparent geometric deduction process.
It can be observed that our method consistently produces significantly more accurate 3D localizations compared to Qwen3-VL-8B-Instruct and VG LLM. Lacking an explicit mechanism to propagate camera intrinsics or depth samples, these models are limited to implicit visual heuristics that inevitably collapse under challenging spatial configurations. In contrast, our equation-anchored framework systematically overcomes this by substituting the explicitly retrieved metric depths and camera intrinsics as deterministic formula variables into the back-projection equation. This strictly anchors the deduced 3D center and dimensions to the true physical scale, yielding 3D bounding boxes that align tightly with the ground-truth objects and demonstrating highly robust, camera-aware spatial comprehension.

\section{Conclusion}

This paper presents an equation-anchored tool-use framework that elevates Multimodal Large Language Models from implicit perceptual estimators to mathematically-grounded spatial agents for precise 3D scene understanding. To overcome the fundamental camera intrinsic ambiguity inherent in single-image 3D localization, we propose a paradigm shift in how spatial tools are utilized. First, our proactive tool invocation mechanism strictly formulates spatial tool outputs, such as multi-point depth samples and camera intrinsics, as explicit formula variables rather than soft reference cues. Second, through a dedicated Chain-of-Thought training pipeline, we establish an explicit geometric reasoning chain that forces the model to seamlessly interleave these numerical variables with transparent symbolic substitutions, actively computing the back-projection equation to deterministically deduce precise 3D coordinates. Extensive experiments demonstrate that our framework effectively resolves scale ambiguity and guarantees camera-robust generalization, achieving highly interpretable and accurate 3D geometric comprehension. Moving forward, we plan to extend this mathematically-grounded reasoning paradigm to dynamic 3D environments to further advance physical-level interactions in embodied AI applications.


\clearpage

\bibliographystyle{plain}
\small\bibliography{reference}

\end{document}